\documentclass[conference]{IEEEtran}
\IEEEoverridecommandlockouts

\usepackage{cite}
\usepackage{amsmath,amssymb,amsfonts}
\usepackage{algorithmic}
\usepackage{graphicx}
\usepackage{textcomp}
\usepackage{xcolor}
\def\BibTeX{{\rm B\kern-.05em{\sc i\kern-.025em b}\kern-.08em
    T\kern-.1667em\lower.7ex\hbox{E}\kern-.125emX}}
    
\begin{document}

\title{Empirical Analysis on Top-k Gradient Sparsification for Distributed Deep Learning in a Supercomputing Environment\\}

\author{\IEEEauthorblockN{Daegun Yoon}
\IEEEauthorblockA{\textit{Department of Artificial Intelligence} \\
\textit{Ajou University}\\
Suwon, Republic of Korea \\
kljp@ajou.ac.kr}
\and
\IEEEauthorblockN{Sangyoon Oh}
\IEEEauthorblockA{\textit{Department of Artificial Intelligence} \\
\textit{Ajou University}\\
Suwon, Republic of Korea \\
syoh@ajou.ac.kr}
}

\maketitle

\begin{abstract}
To train deep learning models faster, distributed training on multiple GPUs is the very popular scheme in recent years. However, the communication bandwidth is still a major bottleneck of training performance. To improve overall training performance, recent works have proposed gradient sparsification methods that reduce the communication traffic significantly. Most of them require gradient sorting to select meaningful gradients such as Top-k gradient sparsification (Top-k SGD). However, Top-k SGD has a limit to increase the speed up overall training performance because gradient sorting is significantly inefficient on GPUs. In this paper, we conduct experiments that show the inefficiency of Top-k SGD and provide the insight of the low performance. Based on observations from our empirical analysis, we plan to yield a high performance gradient sparsification method as a future work.
\end{abstract}


\section{Introduction}
The enormous advance in hardware (e.g., the high performance capability of emerging graphics processing units) accelerates the training time of deep learning models. By building a large-scale cluster equipped with GPUs, it becomes possible to train a very large deep learning model such as GPT-3\cite{gpt3}, Megatron-LM\cite{megatronlm}, and Turing-NLG\cite{turingnlg} in distributed deep learning fashion. However, the advance in communication bandwidth is behind to the computational capability. The low communication bandwidth is a major cause of the performance bottlenecks, and it hinders strong scaling of distributed training even in a large-scale high performance computing system such as a supercomputer. Therefore, thorough analysis on communication optimization methods is critical to improve the performance of distributed training. Moreover, a supercomputing system, which is different from conventional cluster or cloud, is the right platform to conduct empirical analyses to identify important factors related to efficient strong scaling of distributed training.

There are a number of recent research works that identify the communication bottleneck as a major hindrance of the improvement of distributed deep learning training. Among these works, there are proposals of gradient compression methods such as gradient quantization\cite{quantz01, quantz02, quantz03, quantz04} and gradient sparsification\cite{deepgradcomp, adacomp, gaussian, gtopk, lagssgd, omgssgd, scalecom}. From their publications, we can learn that gradient quantization can only compress the gradients in 32$\times$ (i.e., 32 bit to 1 bit at most) because it drops the value under floating point to reduce the size of tensors transmitted through communication. On the other hand, gradient sparsification can reduce the communication traffic over 32$\times$ (e.g., 1000$\times$) because it only sends a few gradients selected by defined policy.

Gradient sparsification aims to reduce the training time at each iteration. However, the actual training time can be longer than the baseline (i.e., distributed deep learning with no adoption of gradient compression) if the optimizer cannot attain the convergence level close to the baseline. To answer this question, several works\cite{convproof01, convproof02} provide theoretical proof of convergence that shows the model with gradient sparsification can attain the convergence level close to the baseline.

\begin{figure}[t]
 \includegraphics[width=1.0\linewidth]{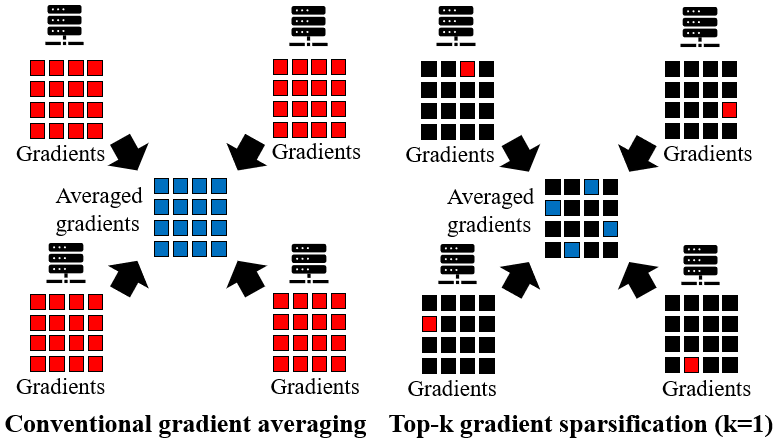}
\caption{Conventional gradient averaging (baseline) and Top-k gradient sparsification (Top-k SGD).}
\label{fig:1}
\end{figure}

Among various gradient sparsification methods, Top-k gradient sparsification, which is called as Top-k SGD in this paper, is the fundamental. Top-K SGD is based on the scheme that selects the $k$ gradients which have the largest values among all gradients. After selection, only $k$ selected gradients are exchanged between workers participating distributed training to calculate globally averaged value of each selected gradient.

Although the convergence of Top-k SGD is proved\cite{convproof02}, its performance suffers from significantly inefficient computation. To determine the indices of Top-k gradients, all gradients of parameters must be sorted, which is generally implemented by torch.topk of PyTorch\cite{pytorch} or tf.math.top\_k of TensorFlow\cite{tensorflow}. Because each GPU core has significantly lower computational capability than a CPU core, sorting tensors which cannot exploit the massive parallelism of streaming multiprocessors is very costly operation on GPUs. Thus, it is worthy to note that compression procedure should not be major bottleneck of overall training performance.

In this paper, we provide empirical comparison results of Top-k SGD with a baseline of distributed training (with no gradient compression) to show the limitation of Top-k SGD due to high computational overhead of gradient sorting. We conducted all of our experiments on a supercomputer, where high quality devices and connection variations (i.e., NVLink and PCI-E) are provided and it is right place to identify how training performance varies with the communication bandwidth.

\begin{figure}[t]
 \includegraphics[width=1.0\linewidth]{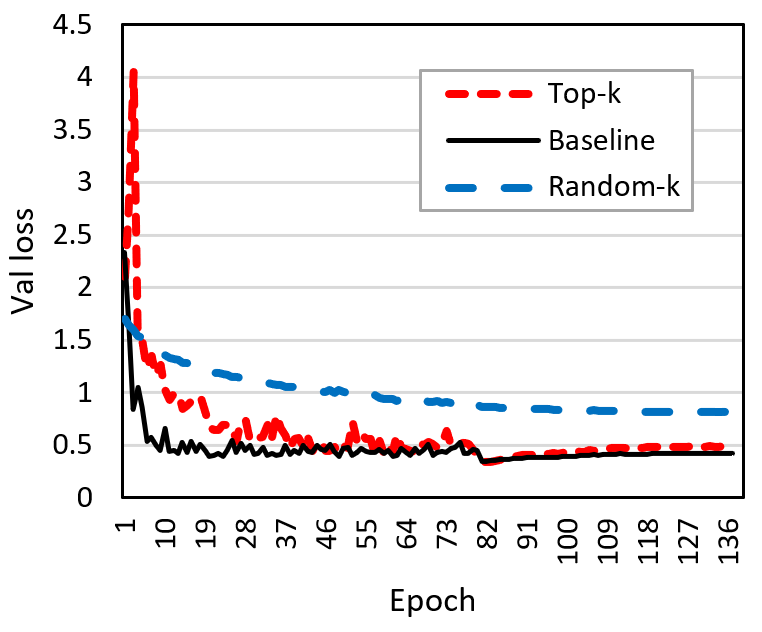}
\caption{Validation loss of Top-k SGD, Random-k SGD, and baseline.}
\label{fig:2}
\end{figure}

\section{Experimental Results}
In this paper, we conducted all experiments using hardware resources provided by the KISTI Neuron\cite{kistineuron} supercomputer where all job scripts are managed by SLURM\cite{slurm}. Basically, we used one of nodes from Neuron that is equipped with eight NVIDIA Ampere A100 GPUs connected with NVLink. For distributed model training, we used PyTorch 1.8.1\cite{pytorch}, Horovod\cite{horovod}, and OpenMPI 3.0.0\cite{openmpi} with CUDA 11.2\cite{cuda}. We used source code of Top-k SGD for training provided by Shi et al\cite{gaussian}, and run it in Python 3.8. Our experiment environment was built using Singularity image (SIF)\cite{singularity} and run in its container. All experiments were conducted with VGG-16 Model and CIFAR-10 dataset. To evaluate the performance of Top-k SGD, Random-k SGD and baseline (i.e., no gradient compression), we set training configuration as follows:
\begin{itemize}
    \item Number of workers = 8
    \item Batch size per worker = 128
    \item Initial learning rate = 0.1
    \item Density: 0.001 for Top-k SGD and Random-k SGD, and 1 for baseline
\end{itemize}

\subsection{Convergence and Accuracy}
In this section, we first compare the convergence and accuracy between Top-k SGD and baseline (with no gradient compression). Fig. 2 depicts the validation loss of Top-k SGD and baseline. Top-k SGD shows slightly higher loss than baseline because Top-k SGD only selects 0.1\% of gradients. Fig. 3 depicts validation top-1 accuracy of Top-k SGD and baseline. Top-k SGD shows slightly lower accuracy than baseline. These experimental results show that Top-k SGD can achieve the convergence performance that is approximate to baseline. 

Figs. 2 and 3 also show the results of Random-k SGD. However, Random-k SGD does not attain the convergence level of other methods in the results. It is because Random-k SGD selects 0.1\% of gradients randomly without any consideration. It shows that the gradient selection of Top-k SGD has a significant meaning which is absent in Random-k SGD.

\begin{figure}[t]
 \includegraphics[width=1.0\linewidth]{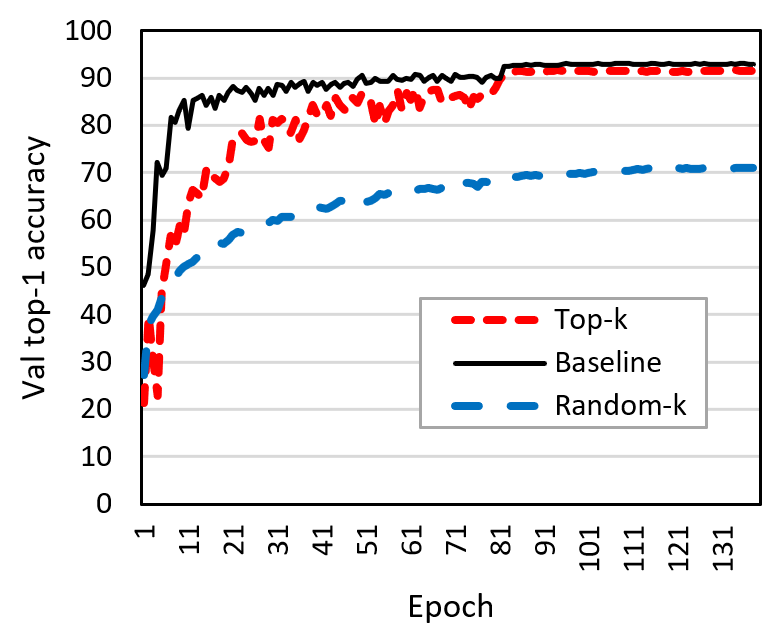}
\caption{Validation top-1 accuracy of Top-k SGD, Random-k SGD, and baseline.}
\label{fig:3}
\end{figure}

\begin{figure}[t]
 \includegraphics[width=1.0\linewidth]{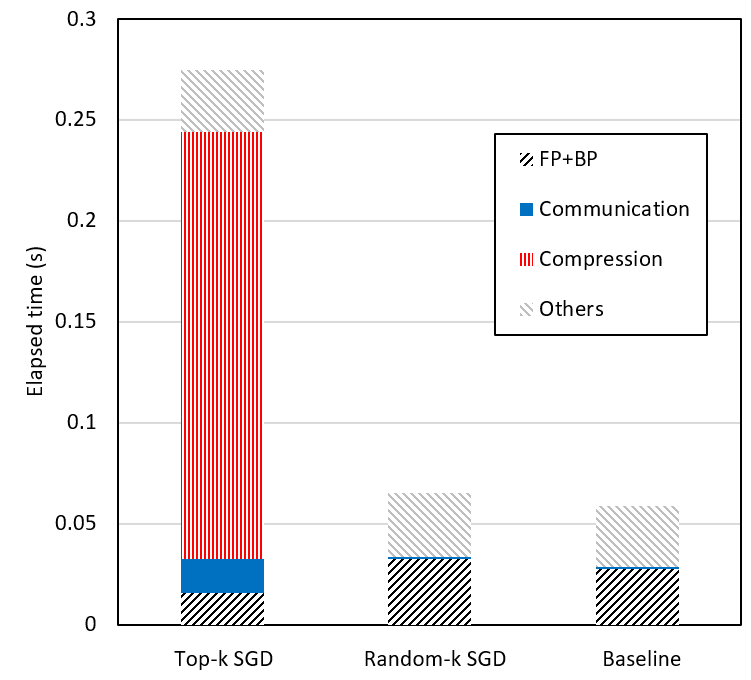}
\caption{Breakdown of iteration. FP and BP mean forward propagation and backpropagation, respectively.}
\label{fig:4}
\end{figure}

\subsection{Breakdown of Iteration}
We measured the training speed for each iteration (i.e., the average of all iterations) of Top-k SGD, Random-k SGD, and baseline. We define the training speed as the number of processed samples per second. The baseline (17387.49 samples/s) achieves the speedup of 4.67$\times$ over Top-k SGD (3723.61 samples/s), and the speedup to Random-k SGD (15648.40 samples/s) is merely and 1.11$\times$. Although Top-k SGD aims to improve training speed for each iteration by reducing communication traffic, the result is presented as opposite. It is because Top-k SGD must sort the all gradients to find $k$ largest gradients.

To dissect this result in detail, we break down the average elapsed time of iterations. Fig. 4 shows the breakdown of each training iteration. The compression time of Top-k SGD is dominant over other criteria. On the other hand, communication occupies small portion of entire runtime. Thus, overall training time of each iteration of Top-k SGD exceeds that of baseline significantly.

The elapsed time for communication of Top-k SGD is also longer than baseline, which is around 16.86ms. It is because allreduce for gradients is implemented as asynchronous. Although gradient sparsification reduces the communication traffic, the time required for synchronization (i.e., synchronize() in Horovod) is longer than that of baseline because the message of several workers may arrive late at that synchronization point due to the long-taken top-k gradient selection. We have an evidence that supports this arguments. The elapsed time for communication in Random-k SGD and baseline are 1.00ms and 1.24ms, respectively. It is because gradient sorting is not included in Random-k SGD. Therefore, gradient sparsification can reduce the communication time when the computational overhead from gradient sorting is resolved.

\subsection{Further Experiments}
Additionally, to analyze the impact of communication bandwidth, we conducted the same experiments with a downgraded network connection from NVLink to PCI-E. However, the training time is reduced in 2\%. From this result, we identify that the gradient sparsification can be used to achieve better performance than baseline when the communication bandwidth is narrow.

To evaluate the performance difference by GPU computational capability, we also conducted experiment using a node with eight NVIDIA Tesla V100 GPUs connected with NVLink. NVIDIA Ampere A100 delivers 312 TFLOPS while NVIDIA Tesla V100 delivers 125 TFLOPS. Except the change of GPU products, other configurations are identical with previous experiments. As the results of experiment in the node with NVIDIA Tesla V100 GPUs, the baseline (12651.83 samples/s) achieves speedup of 3.38$\times$ over Top-k SGD (3743.28 samples/s). The speedup is stayed the same (i.e. not improved) with Top-k SGD when the hardware is upgraded from V100 node to A100 node. On the other hand, baseline achieves speedup of 1.37$\times$ over the result in V100 node. It is because overall performance, which is dominated by gradient sorting, is not improved by upgrading of GPUs. Therefore, it is crucial to use a novel gradient selection method in place of gradient sorting to exploit the computational capability of high quality GPUs.

\section{Conclusion and Future Work}
In this paper, we identify the limitation of Top-k gradient sparsification, which is mainly caused by gradient sorting. Our empirical experimental results support this argument. In experimental results, computational overhead of gradient sorting is dominant over others including communication which occupies small portion. Moreover, we identify the gradient sparsification can be useful to reduce the communication time when communication bandwidth is low.

As future work, we plan to design a computation-efficient gradient sparsification method to resolve computational overhead and delayed synchronization from this overhead, which is different with existing methods that require gradient sorting.

\section*{Acknowledgment}
This work was jointly supported by the Korea Institute of Science and Technology Information (KISTI) (TS-2022-RE-0019), the Basic Science Research Program (2021R1F1A1062779) of the National Research Foundation of Korea (NRF) funded by the Ministry of Education, and the ITRC (Information Technology Research Center) support program (IITP-2022-2018-0-01431) supervised by the Institute for Information \& Communications Technology Planning \& Evaluation (IITP), Korea.


\end{document}